\documentclass[runningheads]{llncs}

\usepackage{hyperref}       % hyperlinks
\usepackage{url}            % simple URL typesetting
\usepackage{booktabs}       % professional-quality tables
\usepackage{amsfonts}       % blackboard math symbols
\usepackage{nicefrac}       % compact symbols for 1/2, etc.
\usepackage{microtype}      % microtypography
\usepackage{lipsum}
\usepackage[dvipsnames]{xcolor}
\usepackage{csquotes}
\usepackage{epsfig}
\usepackage{graphicx}
\usepackage{amsmath}
\usepackage{amssymb}
\usepackage{makecell}
\usepackage{diagbox}

\usepackage{comment}

\usepackage{multirow}
\usepackage{bm}
\usepackage{caption}
\usepackage{amsthm}
\newtheorem{thm}{Theorem}
\newtheorem{defn}{Definition}

\newtheorem{cor}{Corollary}

\newtheorem{prop}{Proposition}

\def\ourmodel{\textit{IRAE}}

\usepackage{bbding}
\usepackage{pifont}
\usepackage{wasysym}
\usepackage{amssymb}
\usepackage{tablefootnote}
\usepackage{authblk}
\usepackage{tabularx} 
\usepackage[misc]{ifsym}

\newcommand{\etal}{\textit{et al}. }
\newcommand{\ie}{\textit{i}.\textit{e}., }

\makeatletter
\newcommand{\printfnsymbol}[1]{%
  \textsuperscript{\@fnsymbol{#1}}%
}
\makeatother

\usepackage[normalem]{ulem}

\newcommand{\x}{\mathbf{x}}
\newcommand{\y}{\mathbf{y}}

\newcommand{\z}{\mathbf{z}}
\newcommand{\m}{\mathbf{m}}
\newcommand{\n}{\mathbf{n}}

\newcommand{\mii}{\mathbb{I}}

\newcommand{\e}{\mathbb{E}}

\usepackage{subcaption}

\begin{document}

%%%%%%%%% TITLE

\title{Are Deep Neural Architectures
\\ Losing Information? \\
Invertibility Is Indispensable
}

\author{Yang Liu \Letter \inst{1,2}\thanks{Equal contribution. \\ 
Code: \texttt{https://github.com/Lillian1082/IRAE\textunderscore pytorch}.  \href{https://github.com/Lillian1082/IRAE_pytorch}{Click Here.} 
} \and
Zhenyue Qin\inst{1}\printfnsymbol{1} \and
Saeed Anwar\inst{1,2} \and
Sabrina Caldwell \inst{1}
\and  \\
Tom Gedeon\inst{1}}

\tocauthor{Yang Liu (ANU & Data61), Zhenyue Qin (ANU), Saeed Anwar (ANU & Data61), Sabrina Caldwell (ANU), Tom Gedeon (ANU)}
\toctitle{Are Deep Neural Architectures Losing Information? Invertibility Is Indispensable}

\authorrunning{Liu \& Qin \etal}
\institute{Australian National University, Australia \and
Data61-CSIRO, Australia 
\\ \email{yang.liu3@anu.edu.au, zhenyue.qin@anu.edu.au, \\saeed.anwar@anu.edu.au,
sabrina.caldwell@anu.edu.au,
tom@cs.anu.edu.au} 
}

\begin{comment}
\author{Anonymous Authors}
\institute{Anonymous Institutions}
\end{comment}
%
\maketitle              % typeset the header of the contribution
%\thispagestyle{empty}

%%%%%%%%% ABSTRACT
\begin{abstract}
   Ever since the advent of AlexNet, designing novel deep neural architectures for different tasks has consistently been a productive research direction. Despite the exceptional performance of various architectures in practice, we study a theoretical question: what is the condition for deep neural architectures to preserve all the information of the input data? Identifying the information lossless condition for deep neural architectures is important, because tasks such as image restoration require keep the detailed information of the input data as much as possible. Using the definition of mutual information, we show that: a deep neural architecture can preserve maximum details about the given data if and only if the architecture is invertible. We verify the advantages of our Invertible Restoring Autoencoder (IRAE) network by comparing it with competitive models on three perturbed image restoration tasks: image denoising, JPEG image decompression and image inpainting. Experimental results show that IRAE consistently outperforms non-invertible ones. Our model even contains far fewer parameters. Thus, it may be worthwhile to try replacing standard components of deep neural architectures with their invertible counterparts. We believe our work provides a unique perspective and direction for future deep learning research. 

\end{abstract}

\section{Introduction}
Ever since AlexNet won the ImageNet challenge in 2012~\cite{krizhevsky2012imagenet}, deep learning has been revolutionizing research in many industries. One key factor to account for the success of deep learning is the transferability of deep learning architectures~\cite{bengio2012deep}. That is, a neural architecture that exhibits excellent performance for one task can also excel in a variety of other related tasks. However, the majority of deep learning architectures were initially proposed to address high-level computer vision tasks. Recently, researchers also explored applying these deep architectures used for high-level tasks to tackle low-level vision tasks. 

Empirical results suggest the plausibility of transferring the neural architectures for high-level vision tasks to addressing low-level image-processing tasks. Nonetheless, there is a division between the requirements for deep models to solve high- and low-level vision tasks. To specify, it may be acceptable to miss image details for high-level tasks, as long as it captures the most salient features. However, missing details of images can be unsupportable when dealing with low-level vision tasks. Instead of primarily concentrating on conceptual vision features, models for low-level tasks require specific minutiae such as colors and textures to be able to restore original images~\cite{liu2020gradnet}. 

Inspired by the division between the requirements for high- and low-level vision tasks, we study whether it is proper to apply deep architectures of high-level vision tasks to tackle low-level tasks. From the perspective of mutual information, we show that: in order to let a neural architecture to preserve all the information of the given input, the neural architecture needs to be invertible. In this paper, we evaluate the performance of invertible neural architectures on image restoration tasks. Invertible neural architectures exhibit excellence experimental results. Thus, we believe it is a promising avenue to replace non-invertible neural components with their invertible counterparts. In summary, our contributions are three-fold: 
\vspace{-2.05mm}
\begin{enumerate}
    \item Deriving from the definition of mutual information, we show non-invertible deep neural architectures lead to loss of information concerning the input. 
    \item Inspired by the need for invertibility, we develop an Invertible Restoring Autoencoder (IRAE) network via invertible flow-based generative algorithms.  
    \item We test IRAE with a series of experiments, finding that we achieve superior performance on both image denoising and inpainting tasks. Moreover, our model has fewer parameters than the baseline information-lossy models. 
\end{enumerate}

\section{Preliminaries}

\subsection{Residual Blocks}
Deep neural networks suffer from the problem of vanishing gradient~\cite{hochreiter1998vanishing} when the depth of the network increases. To address this problem, Residual Networks and Highway Networks~\cite{srivastava2015highway} use additional pathways to connect the input with the output of a layer directly. Such residual paths facilitate back-propagation, bypassing the multiplication with the layer weight to alleviate the vanishing gradient. These residual blocks are common features in image restoration models. For example, REDNet~\cite{mao2016image} uses symmetric residual blocks. Zhang~\etal \cite{zhang2020residual} employ a large number of residual blocks to preserve detailed information. 

\subsection{Flow-Based Generative Models}
\label{subsec:flow}
The arguably most important cornerstone of generative models is maximum likelihood estimation (MLE). Generative models aim to maximize the probabilities of producing results that look similar to the given data. Unlike variational autoencoders (VAE) \cite{kingma2013auto} and generative adversarial networks (GANs) \cite{goodfellow2014generative} that bypass accurately estimating densities, flow-based generative models directly maximize log probabilities of the given data. Therefore, flow-based generative models require all the model components to be invertible. Pioneering flow-based generative models include NICE \cite{dinh2014nice} and RealNVP \cite{dinh2016density}. However, they suffer from poor generation quality. Recently, Glow was proposed by Kingma \etal \cite{kingma2018glow}. Glow can generate realistic-looking images, achieving similar and sometimes better performance than other generative algorithms like VAE and GAN. 

\subsection{Mutual Information}
Mutual Information (MI) is a quantity measuring the dependencies between two variables. Intuitively, it estimates the amount of information that one can obtain about one variable when observing the other~\cite{belghazi2018mut}. MI is more powerful than correlations. Correlations can only measure dependencies between two linearly dependent variables. MI can tackle non-linearity among variables~\cite{qin2019rethinking}. Thus, MI is employed to investigate how learning is achieved in a deep neural network with many non-linear layers. Examples include pc-softmax and the information-bottleneck theory~\cite{qin2019rethinking,saxe2019information}. The formula for MI $\mii(\x; \y)$ between variables $\x$ and $\y$ is:
\begin{gather*}
    \mii(\x; \y) = \e_{(\x,\y)} \left[ \log \left(\frac{P(\x, \y)}{P(\x) P(\y)} \right)\right]. 
\end{gather*}

\section{Conditions when Deep Architectures Lose Information}
\label{sec:cond_lose_info}
In this section, we show interesting circumstances under which a deep neural architecture loses information about the given input data. To this end, we first present the definition of an invertible deep neural architecture: 
\begin{defn}
\label{defn:invertible_arch}
A deep neural architecture is invertible if and only if: 
\begin{enumerate}
    \item It satisfies the function property of being deterministic. 
    \item It meets the definition of a one-to-one function. 
\end{enumerate}
\end{defn}
Each input $\x$ corresponds to a unique resultant variable $\z$, We consider the cases of the variables being discrete and continuous separately. If the variables are discrete, one immediate implication of \autoref{defn:invertible_arch} is $P(\x | \z) = 1$, \ie the conditional probability of the input data $\x$ given the output $\z$ is one, where the output $\z$ can either be the intermediate features or the final output. In contrast, for discrete variables, a non-invertible deep neural architecture has $P(\x | \z) < 1$, since multiple different input $\x$ can lead to the same output $\z$. Furthermore, if the variables are discrete, we consider $P(\x | \z)$ as a probability, thus it cannot exceed 1. As to the cases of continuous variables, we use a conclusion from~\cite{kraskov2004estimating}, stating that MI is invariant under the smooth invertible transformations of the variables, to show invertible can preserve information about the input data. We then have the following proposition: 
\begin{prop}
\label{prop:invertible}
If a deep neural architecture is not invertible, then it will lose information from the input during the feed-forward process. 
\end{prop}
\vspace{-3mm}
\label{thm:nn_inv}

\begin{proof}
Following the above definition, we use $\x$ and $\z$ to respectively denote the input data and the middle or final features of $\x$ processed by a neural network. We employ MI, denoted as $\mii(\x; \z)$, to represent the information that $\z$ carries about $\x$. From the definition of MI, we have: 
\begin{gather*}
    \mii(\x; \z) = \e_{(\x,\z)} \left[ \log \left(\frac{P(\x, \z)}{P(\x) P(\z)} \right) \right]
    = \e_{(\x,\z)} \left[ \log \left(\frac{P(\x | \z)}{P(\x)} \right) \right]. 
\end{gather*}
From the definition of invertible deep neural architectures, if variables $\x$ and $\z$ are discrete, we have $p(\x | \z) = 1$ if and only if the architecture is invertible. In contrast, when not invertible, we have: $p(\x | \z) < 1$. That is, the MI between $\x$ and $\z$ is larger when the network is invertible as compared to not being invertible. 

When variables $\x$ and $\z$ are continuous, we utilize the fact that MI is invariant under the smooth invertible transformations of the variables~\cite{kraskov2004estimating}. That is, given two continuous variables $\m$ and $\n$, we have: 
\begin{equation}
    \mii(\m; \n) = \mii(f(\m); g(\n)), 
\end{equation}
where both functions $f$ and $g$ are smooth and invertible functions. Therefore, if function $f(\x) = \x$, and function $g(\x)$ is an invertible neural architecture, outputting $\z$, \ie $g(\x) = \z$. Then, we have: 
\begin{equation}
    \mii(\x; \x) = \mii(f(\x); g(\x)) = \mii(\x; \z). 
\end{equation}
That is, the MI between input $\x$ and output $\z$ is the same as the MI between input $\x$ and itself, implying output $\z$ preserves information of input $\x$. 
\end{proof}

\begin{comment}
\textcolor{red}{Clearly specify, when $p(\x | \z) < 1$ and $p(\x | \z) > 1$. There is a small confusion in the above paragraph.}
\end{comment}

\begin{comment}
Following Theorem 1, we have the following corollary: 
\begin{cor}
\label{cor:resnet}
Non-invertible components of deep neural architectures such as residual blocks and ReLU cause information loss. 
\end{cor}
\end{comment}

From \autoref{prop:invertible}, we note that we require invertible deep neural networks to maintain all the detailed information about the input data. 

\begin{comment}
\section{Conditions of Deep Architectures Losing Information}
In this section, we show under what circumstances a deep neural architecture can lose information about the input data. 
\begin{thm}
If a deep neural network is not invertible, then it will lose information about the input during the feedforward process. 
\end{thm}
\label{thm:nn_inv}
\begin{proof}
We use $\x$ and $\z$ to respectively denote the input data and the middle feature of $\x$ processed by a neural network. We employ mutual information (MI) $\mii(\x; \z)$ to represent the information that $\z$ carries about $\x$. From the definition of MI, we have: 
\begin{gather*}
    \mii(\x; \z) = \e_{(\x,\z)} \big[ \log \frac{p(\x, \z)}{p(\x) p(\z)} \big]
    = \e_{(\x,\z)} \big[ \log \frac{p(\x | \z)}{p(\x)} \big]. 
\end{gather*}
We have $p(\x | \z) = 1$ if and only if when the neural network from $\x$ to $\z$ are invertible. In contrast, when not invertible, we have: $p(\x | \z) < 1$. That is, the MI between $\x$ and $\z$ is larger when the network is partially invertible from $\x$ to $\z$ than not being invertible. 
\end{proof}

Following \autoref{thm:nn_inv}, we have the following corollary: 
\begin{cor}
\label{cor:resnet}
Non-invertible components of deep neural architectures such as residual blocks and ReLU cause information loss. 
\end{cor}

From \autoref{thm:nn_inv} and \autoref{cor:resnet}, we observe that one requires a deep neural networks that are invertible in order to maintain all the detailed information about the input data. 
\end{comment}
\section{Flow-Based Image Restoration Models}
\label{section:flow_based_model}

\begin{figure}[t]
    \centering
    \includegraphics[width=\linewidth]{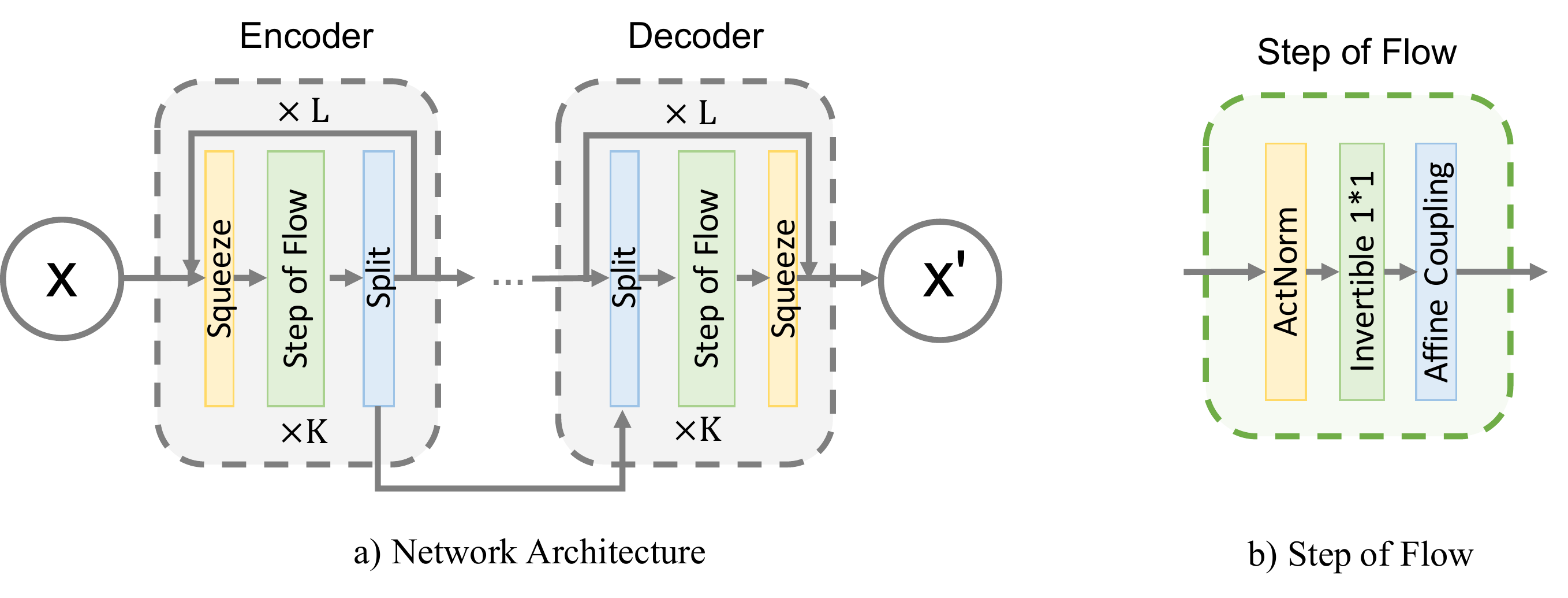}
    \caption{Architecture of our Invertible Restoring Autoencoder (IRAE) network. \enquote{ActNorm} means \enquote{activation normalization}.}
    \label{fig:my_label}
    \vspace{-6mm}
\end{figure}

In \autoref{subsec:flow}, we have described the requirement of flow-based generative models. Flow-based generative models require an invertible mapping between the input and the latent tensors, directly conducting maximum likelihood estimation for the given data. Due to the invertibility between inputs and outputs, as \autoref{prop:invertible} suggests, flow-based models are information-lossless. 

We also require a image-restoration model to be information-lossless. Therefore, we investigate the empirical performance of applying flow-based invertible deep architectures to address image restoration tasks.  

% \vspace{-4mm}
\subsection{Architecture Overview}
% \vspace{-2mm}
Figure~\ref{fig:my_label} presents an overview of our deep architecture for image restoration. Our primary requirement is to make the architecture completely invertible to preserve all information about the given data, as we have described in \autoref{sec:cond_lose_info}. To fulfill the invertible requirement, we aim for an encoding-decoding symmetric image-restoration deep architecture. In the architecture, every component is invertible. In the subsequent sections, we describe each component of our architecture in detail.

% \vspace{-4mm}
\subsection{Encoder and Decoder}
% \vspace{-2mm}
Our auto-encoding architecture is inspired by Glow~\cite{kingma2018glow}, a flow-based generative model. Two motivations inform this choice: 1) We require an invertible deep architecture to preserve all the information about the input data, and 2) among flow-based models, Glow can generate the highest quality images. However, our architecture is considerably different from the Glow architecture. While Glow contains only an encoder and relies on the inverse function of the encoder to reconstruct the images, we utilize an additional decoder, and hence our architecture is symmetric, as Figure~\ref{fig:my_label} depicts. We argue that the decoder is mandatory because, unlike reconstructing the given images, our model aims to convert the given perturbed data to the original forms. 
\begin{comment}
The exact inversion cannot address this restoration \textcolor{red}{what this sentence means}.
\end{comment}
Nonetheless, the entire architecture is still invertible since encoders and decoders are invertible. 
% , and the intermediate transformation component 
\begin{comment}
Our autoencoding architecture is inspired from a flow-based generative model coined Glow. Two motivations support this choice: 1. We require an invertible deep architecture to preserve all the information about the input data. Flow-based models can satisfy this requirement. 2. Among flow-based models, Glow can generate the highest quality images. However, we are far different from copying the exact Glow architecture. Unlike Glow containing only an encoder and relies on the inverse function of the encoder to reconstruct the images, we demand an additional decoder so our architecture is of symmetry as Figure x depicts. The decoder is mandatory because unlike reconstructing the given images, our model aims to convert the given perturbed data to the original forms. The exact inversion cannot address this restoration. Nonetheless, the entire architecture is still invertible since encoders, decoders and the intermediate transformation component are all invertible. 
\end{comment}

\subsection{Invertible Local Spatial Feature Extraction}
The great success of CNNs can be attributed to their ability to leverage local spatial features. Specifically, CNN filters can exploit the 2D-spatial structure of images via employing spatial convolutions to extract the local information around each pixel. Unfortunately, this operation is non-invertible due to dimension reduction, which leads to further information loss. To address this dimension reduction, we borrow ideas from a flow-based model called Real-NVP and utilize the spatial checkerboard pattern within Real-NVP to apply the local spatial feature aggregation. Particularly, we squeeze a $1 \times 4 \times 4 $ tensor, where the order is \enquote{channel $\times$ width $\times$ height}, into a $4 \times 2 \times 2$ tensor. Consequently, each resultant channel corresponds to a $4 \times 4$ region of the original image. We then perform a $1 \times 1$ convolution to aggregate channel information together. In this squeezing and $1 \times 1$ convolving fashion, we facilitate extracting local spatial features invertibly. 

\subsection{Steps of Flow}
We follow the same set of flow steps as the Glow model, consisting of three invertible sub-steps:
\begin{enumerate}
    \item \textbf{Activation Normalization} abbreviated as ActNorm, performs an affine transformation with a scale and a bias parameter per channel. The intention is to initialize the first minibatch to have mean zero and standard deviation of one after ActNorm to address covariate shift, similar to batch normalization. 
    \item \textbf{$1 \times 1$ Convolution}  can be viewed as a linear transformation without shrinking dimensions. Thus, it is invertible. 
    \item \textbf{Affine Coupling} aims to mix information of different dimensions. It consists solely of invertible functions, so the composite function is still invertible. 
\end{enumerate}

\subsection{Loss Function}
Assume the training pairs we use are $\{\bm{x_i}, \bm{y_i}\}_{i=1}^N$, where $\bm{x_i}$ is the ground truth image and $\bm{y_i}$ is the corresponding corrupted image,
\begin{equation}
    \bm{y_i} = \bm{A}\otimes\bm{x_i} + \bm{n_i},
\end{equation}
where $\otimes$ is the element-wise multiplication. $\bm{A}$ is the degradation matrix (which is an identity matrix for image denoising tasks).
We follow the standard loss function for image restoration tasks and use the $\ell_1$ loss as the objective function,

% Assume the training pairs we use are $\{\bm{x_i}, \bm{y_i}\}_{i=1}^N$. $\bm{x_i}$ is the ground truth image and $\bm{y_i}$ is the corresponding corrupted image.
% \begin{equation}
%     \bm{y_i} = \bm{A}\otimes\bm{x_i} + \bm{n_i}
% \end{equation}
% where $\otimes$ is the element\-wise product. $\bm{A}$ is the degradation matrix which is an identity matrix for image denoising tasks.
% We follow the standard loss function for image restoration tasks and use the $\ell_1$ loss as the objective.

% Assume the input-output training pairs of our network are $\{\bm{x_i}, \bm{y_i}\}_{i=1}^N$ which are related to each other through $\bm{y_i} = \bm{x_i} + \bm{n_i}$. $\bm{x_i}$ is the clean image and $\bm{y_i}$ is the corresponding noisy image. Let $\hat{\bm{x_i}}$ to be the naive denoised image and $\widetilde{\bm{x_i}}$ to be the output of our network. We use the $\ell_1$ loss as the objective of the backbone network, $\text{GradNet}(\cdot)$,
\begin{equation} \label{equ: objective}
%\begin{split}
     %&min:  
     L_p = \frac{1}{N}\sum_{i=1}^N || \text{\ourmodel}(\bm{y_i}) - \bm{x_i} ||_1 ,\\
    %  &min:  L(W) =  L_p+ \lambda* L_g(\bm{\widetilde{x}_i})),
     %\end{split}
\end{equation}
where $L_p$ is the pixel loss and Invertible Restoring Autoencoder (\ourmodel) is our network.
\begin{comment}
\textcolor{red}{Give a name to the network and use that instead of $NN(\cdot)$}
\end{comment}
\section{Experiments}
To evaluate the image restoration performance of our invertible deep neural architecture, we conduct experiments on three tasks: 1) image denoising, 2) JPEG compression, and 3) image inpainting. All of these experiments show that our model can consistently restore images to their original forms, and better than other competitive methods. The quantitative results of noise removal even show a large-margin improvement. More pleasantly, despite superior performance, our invertible architecture contains fewer parameters than other competitive models. Specifically, our model has $1.33 \times 10^{6}$ parameters, whilst DnCNN \cite{zhang2017beyond} and U-Net~\cite{ronneberger2015u} have $1.48 \times 10^6$ and $7.70 \times 10^6$ parameters respectively. 
\begin{comment}
We conduct three experiments to evaluate the image restoration performance of our invertible deep neural architectures: synthetic noise removal, real-image denoising and image inpainting. Each of these experiments show that our model can consistently restore images to their original forms better than other competitive baseline methods. The quantitative results of synthetic noise removal even show a large-margin improvement. Importantly, despite a more superior performance, our invertible architecture contains fewer parameters than other competitive models. Specifically, our model has $1.33 \times 10^{6}$ parameters, whilst DnCNN \cite{zhang2017beyond} and U-Net have $1.48 \times 10^6$ and $7.70 \times 10^6$ parameters respectively. 
\end{comment}

\subsection{Experimental Settings}
For our architecture \ourmodel~(see Figure~\ref{fig:my_label}) (Invertible Restoring Autoencoder), we apply $K = 16$ and $L = 2$ for the encoding and decoding layers. We adopt Adam \cite{kingma2014adam} as the optimizer and the learning rate as $10^{-3}$ initially, which decays to $2\times10^{-4}$ after the first 50 epochs. Afterward, if the peak signal-to-noise ratio (PSNR) does not improve for 10 epochs, the learning rate decays to a fifth of the original rate. The training terminates when the learning rate decreases below $10^{-6}$. 
% and $B = 5$ for transformation blocks

The evaluation metrics we use for comparison are the average peak signal-to-noise ratio (PSNR) and the structural similarity index (SSIM). Higher values indicate better performance for both metrics. 

\begin{comment}
For the architecture of InvIR as depicted in Figure x, we apply $K = 16$ and $L = 2$ for the encoding and decoding layers and $B = 5$ for transformation blocks. We adopt Adam as the optimizer. The learning rate is initialized as $10^{-3}$ and decays to $2\times10^{-4}$ after the first 50 epochs. The learning rate decay afterwards is to become a fifth of the original rate if the PSNR does not improve for 10 epochs. We terminate training when the learning rate becomes less than $10^{-6}$.
\end{comment}

% \subsection{Metrics}

\begin{table*}[t] %\small
\centering
\bgroup
\setlength{\tabcolsep}{0.0em}
\begin{tabular}{l@{\quad} |@{\quad} c @{\quad} | @{\quad} c @{\quad} | @{\quad} c @{\quad} | @{\quad} c @{\quad} |@{\quad} c @{\quad}}
\toprule
Dataset & $\sigma$ & DnCNN \cite{zhang2017beyond} & FFDNet\cite{zhang2018ffdnet} & U-Net \cite{ronneberger2015u} & \textbf{Ours} \\ \hline

\multirow{4}{3em}{CelebA} & 15  & 31.4597 & 31.8551 & 31.7719 & \textbf{33.0812} \\ 
& 25 & 29.4340 & 28.9438 & 30.5915 & \textbf{31.0795} \\ 
& 50 & 26.7126 & 25.5980 & 26.4775 & \textbf{28.0887} \\ 
& BLD & 30.3496 & 30.8492 & 30.8807 & \textbf{31.6158} \\
\hline
\multirow{4}{3em}{Bird} & 15 & 30.0825 & 31.1383 & 31.9578 & \textbf{33.0805} \\
& 25 & 28.4882 & 28.3442 & 30.3652 & \textbf{31.0111} \\ 
& 50 & 26.4529 & 26.4082 & 27.7812 & \textbf{28.0976} \\ 
& BLD & 28.7662 & 28.4686 & 31.4173 & \textbf{31.4448} \\
\hline
\multirow{4}{3em}{Flower} & 15 & 29.8116 & 30.8773 & 32.0690 & \textbf{32.6267} \\
& 25 & 28.8444 & 28.2578 & 30.2644 & \textbf{30.5044} \\ 
& 50 & 26.1073 & 26.1131 & 27.1729 & \textbf{27.5092} \\ 
& BLD & 28.4645 & 28.7165 & 31.3378 & \textbf{31.6123} \\
\bottomrule
\end{tabular}
\egroup
\caption{Quantitative denoising performance of our model compared against others. The column $\sigma$ stands for different noise levels. \enquote{BLD} represents blind denoising. Values are average PSNR(dB). The best results are highlighted in bold. }
\label{Tab:Synnthetic Denoising}
\vspace{-10mm}
\end{table*}

\begin{figure}
    \centering
    \includegraphics[width=\linewidth]{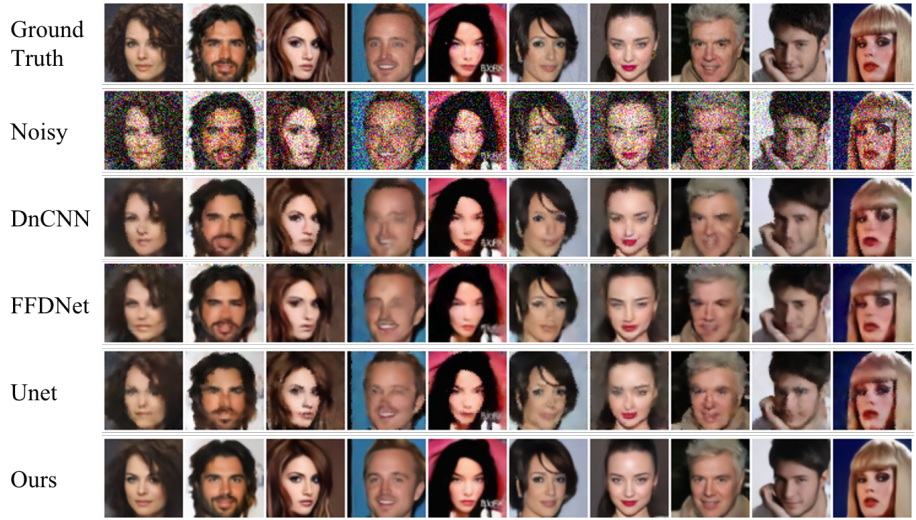}
    \caption{Qualitative visualization of image denoising of our model compared with other methods. The noise level $\sigma = 50$. }
    \label{fig:Denoising}
\end{figure}

\begin{table*}[t] %\small

\centering
\begin{subtable}[t]{\linewidth}
\centering
\bgroup
\setlength{\tabcolsep}{0.0em}
\begin{tabular}{l@{\quad} |@{\quad} c @{\quad} | @{\quad} c @{\quad}| @{\quad} c @{\quad}|@{\quad} c @{\quad}}
\toprule
\diagbox{QF}{Mod} & 
AR-CNN\cite{dong2015compression} & DnCNN\cite{zhang2017beyond} & U-Net\cite{ronneberger2015u} & \textbf{Ours} \\ \hline 
10 & 27.8221 & 27.5406 & 28.7041 & \textbf{28.7322} \\ 
20 & 29.9814 & 28.8870 & 30.9274 & \textbf{30.9984} \\
30 & 31.1935 & 28.6428 & 32.0025 & \textbf{32.1832} \\
40 & 31.9283 & 31.0741 & 32.8755 & \textbf{33.0835} \\
\bottomrule
\end{tabular}
\caption{Quantitative JPEG decompression results in PSNR (dB). }
\egroup
\end{subtable}

\begin{subtable}[t]{\linewidth}
\centering
\bgroup
\setlength{\tabcolsep}{0.0em}
\begin{tabular}{l@{\quad} |@{\quad} c @{\quad} | @{\quad} c @{\quad}| @{\quad} c @{\quad}|@{\quad} c @{\quad}}
\toprule
\diagbox{QF}{Mod} & 
AR-CNN\cite{dong2015compression} & DnCNN\cite{zhang2017beyond} & U-Net\cite{ronneberger2015u} & \textbf{Ours} \\ \hline 
10 & 0.9368 & 0.9326 & \textbf{0.9479} & 0.9476 \\ 
20 & 0.9568 & 0.9469 & 0.9647 & \textbf{0.9652} \\
30 & 0.9658 & 0.9350 & 0.9712 & \textbf{0.9715} \\
40 & 0.9702 & 0.9648 & 0.9757 & \textbf{0.9765} \\
\bottomrule
\end{tabular}
\caption{Quantitative JPEG decompression results in SSIM. }
\egroup
\end{subtable}

\caption{Quantitative JPEG decompression performance of our model compared against others. 'QF' means \enquote{quality factor}. Higher QFs represent less compression loss. PSNR values are average (in dBs). The best results are highlighted in bold. }
\label{table:jpeg_decomression}

\end{table*}

\begin{figure}
    \centering
    \includegraphics[width=0.85\linewidth]{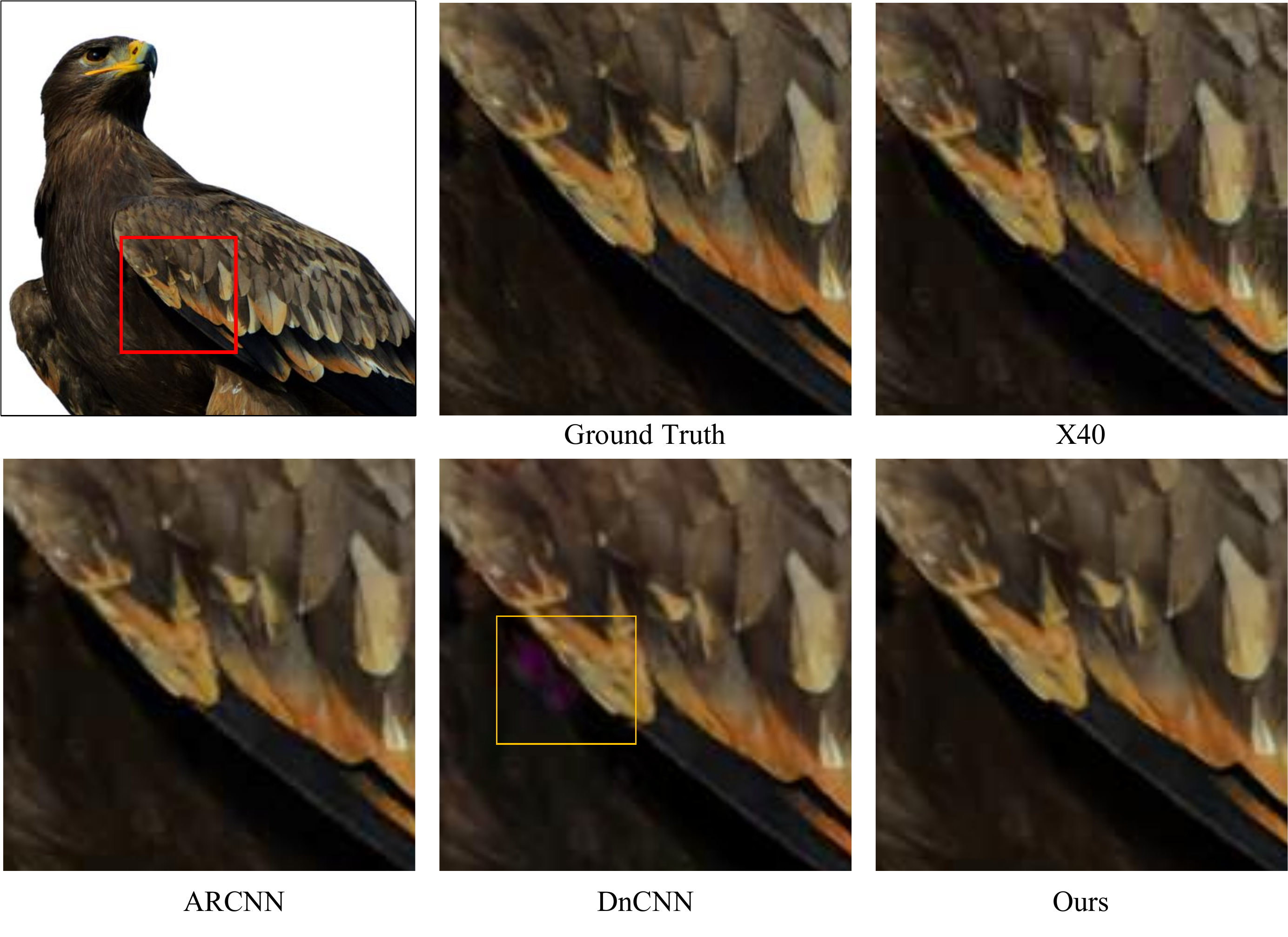}
    \caption{Qualitative visualization of JPEG decompression of our model compared with other methods. Our model preserves details more clearly and does not have artifacts as in the yellow box in DnCNN while remaining invertible and requiring few parameters. }
    \label{fig:jpeg_decomression}
\end{figure}

\begin{table*}[t] %\small
\centering
\bgroup
\setlength{\tabcolsep}{0.0em}
\begin{tabular}{l@{\quad} |@{\quad} c @{\quad} | @{\quad} c @{\quad}| @{\quad} c @{\quad}|@{\quad} c @{\quad}}
\toprule
\diagbox{Metrics}{Models} & \makecell{Contextual \\ Attention \cite{yu2018generative}} & Shift-Net \cite{yan2018shift} & \makecell{Coherent \\ Semantic \cite{liu2019coherent}} & \textbf{Ours}  \\ \hline 
PSNR & 23.93 & 26.38 & 26.54 & \textbf{27.14}  \\ \hline
SSIM & 0.882 & 0.926 & 0.931 & \textbf{0.975} \\
\bottomrule
\end{tabular}
\egroup
\caption{Quantitative results of our model compared with others on image inpainting, in terms of PSNR (averaged) on CelebA dataset. }
\label{Tab:Center_holes}
\vspace{-6mm}
\end{table*}

\begin{figure}
    \includegraphics[width=0.9\linewidth]{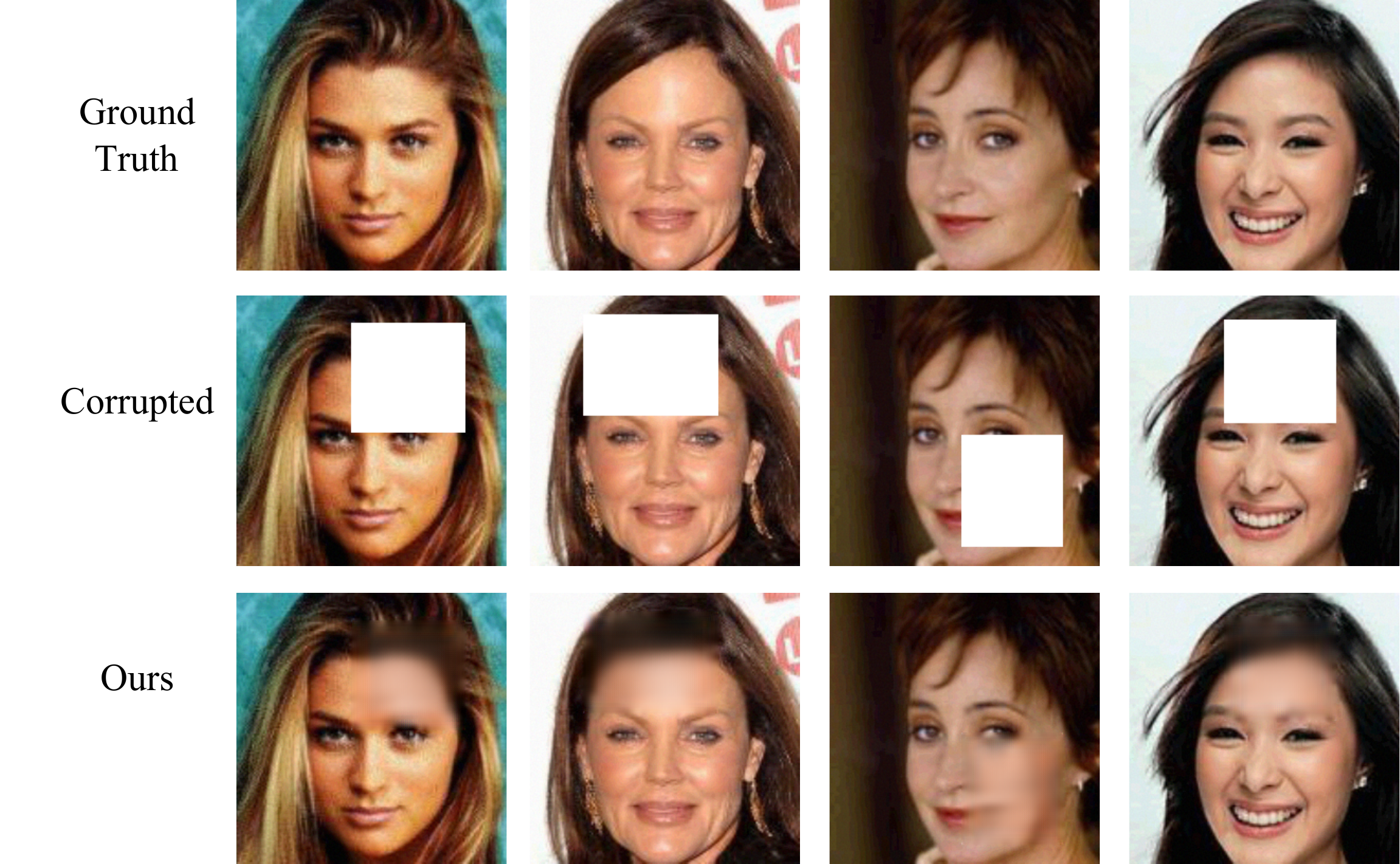}
    \caption{Image inpainting. Our model is general and has not been customized for image inpainting like the state-of-the-art methods.}
    \label{fig:inpainting}
    \vspace{-2mm}
\end{figure}

\subsection{Comparison on Image restoration tasks}
To show the competitive performance of our \ourmodel~ model, we compare its performance on the following three image restoration tasks.

\begin{itemize}
\item \textbf{Image Noise Removal.}
We first evaluate the denoising capability of \ourmodel~ on three types of images: CelebA (human faces)~\cite{liu2018large}, Flower (natural plants)~\cite{nilsback2008automated} and Bird~\cite{wah2011caltech}. The synthetic noise added to these datasets is standard additive white Gaussian noise (AWGN) \cite{thangaraj2017capacity}, with three distinct standard deviations: $\sigma=15, 25, 50$. We also conduct experiments for blind denoising with random noise levels between 0 to 55. \autoref{Tab:Synnthetic Denoising} quantitatively demonstrate that our model performs denoising better than other approaches by a large margin. \autoref{fig:Denoising} exhibits qualitative results of the denoising performance of our model compared against others. 

\vspace{2mm}
\item  \textbf{JPEG Image Decompression.}
JPEG is a commonly used lossy image compression method. It achieves compression by converting images into a frequency domain, then discarding the high-frequency regions that are hard to perceive for humans \cite{liu2019downscaling}. However, JPEG compression often leads to artefacts, such as blockiness and 'mosquito noise'. We evaluate the capability of \ourmodel~ to decompress JPEG images in comparison with competitive methods. Our model can reconstruct JPEG images back to their near-original forms. The effectiveness is shown in  \autoref{table:jpeg_decomression} for quantitative results and \autoref{fig:jpeg_decomression} for qualitative visualization. 
As illustrated in Table 2, our architecture achieves the highest PSNR results on the images with different compression quality factors. Although Unet also gets competitive results, the parameter in our network is $1.33 \times 10^{6}$, which is far less than the parameter used in Unet ($7.70 \times 10^{6}$). For factor 10, we achieved almost the same SSIM result but higher PSNR value than Unet, which also demonstrates our method's superiority in denoising.

\vspace{2mm}
\item  \textbf{Image Inpainting.}
Lastly, we show that our invertible deep neural architecture also performs better for image inpainting. We employ the CelebA dataset with a size of $256 \times 256$, followed by randomly generating masks of size $128 \times 128$, overlapped on each image. We ignore masking parts that are outside the central region of images. We are pleased to find that our model outperforms, even by a large margin, other methods that are specifically designed for image inpainting, such as the ones based on adversarial training, as \autoref{Tab:Center_holes} indicates. \autoref{fig:inpainting} presents a qualitative visualization of our model's results on image inpainting.
\end{itemize}

\section{Discussion and Future Work}

One may concern that: preserving all the information of the given images is not logically plausible. To specify, the noises on an image is also a part of the information. Then, how can we conduct denoising if we also preserve the information of noises? However, preserving the noise information is not equivalent to keeping noises visible. That is, if a model maps noises to values that are extremely close to zero, then noises on images become invisible. Consequently, the model performs denoising well. Also, from the information-theoretic perspective, the model preserves all the information of the given image. Nevertheless, if the model loses the information of the visually salient regions, then even the model can remove all the noises, it is still not acceptable. 

This paper can be inception of investigating whether performance improves if neural network components become invertible. With the definition of mutual information, we have shown the necessity of invertibility for preserving information. We have also demonstrated promising results of using invertible neural networks for image restoration. 

\section{Conclusion}
Designing deep neural architectures is an essential role in modern deep learning research. In the past decade, the manually designed deep neural architectures such as VGGs~\cite{simonyan2014very} and GoogLeNet~\cite{szegedy2015going} have made breakthroughs in various applications. Recently, automatic searching for effective deep neural architectures has gained attention~\cite{elsken2019neural}. In this paper, we aim to study a theoretical question: what deep neural architectures can preserve all the information of the input data? We leverage the definition of mutual information. We show that: invertible deep neural architectures are indispensable to preserve all the details about the given data. We propose IRAE, an invertible model. Experimental results of IRAE for image denoising, decompressing, and inpainting further validate the necessity of invertibility. Our IRAE even has far fewer parameters. We believe our theoretical results and practical demonstration in this paper imply that: making deep neural architectures invertible can be a promising future direction for deep learning research.

% \clearpage
{\small
\bibliographystyle{splncs04}
% \bibliography{egbib}

\begin{thebibliography}{10}
\providecommand{\url}[1]{\texttt{#1}}
\providecommand{\urlprefix}{URL }
\providecommand{\doi}[1]{https://doi.org/#1}

\bibitem{belghazi2018mut}
Belghazi, M.I., Baratin, A., Rajeshwar, S., Ozair, S., Bengio, Y., Courville,
  A., Hjelm, D.: Mutual information neural estimation. In: International
  Conference on Machine Learning. pp. 531--540 (2018)

\bibitem{bengio2012deep}
Bengio, Y.: Deep learning of representations for unsupervised and transfer
  learning. In: Proceedings of ICML workshop on unsupervised and transfer
  learning. pp. 17--36 (2012)

\bibitem{dinh2014nice}
Dinh, L., Krueger, D., Bengio, Y.: Nice: Non-linear independent components
  estimation. ICLR  (2015)

\bibitem{dinh2016density}
Dinh, L., Sohl-Dickstein, J., Bengio, S.: Density estimation using real nvp.
  In: International Conference on Machine Learning (2016)

\bibitem{dong2015compression}
Dong, C., Deng, Y., Change~Loy, C., Tang, X.: Compression artifacts reduction
  by a deep convolutional network. In: Proceedings of the IEEE International
  Conference on Computer Vision. pp. 576--584 (2015)

\bibitem{elsken2019neural}
Elsken, T., Metzen, J.H., Hutter, F.: Neural architecture search: A survey.
  Journal of Machine Learning Research  \textbf{20},  1--21 (2019)

\bibitem{goodfellow2014generative}
Goodfellow, I., Pouget-Abadie, J., Mirza, M., Xu, B., Warde-Farley, D., Ozair,
  S., Courville, A., Bengio, Y.: Generative adversarial nets. In: Advances in
  neural information processing systems. pp. 2672--2680 (2014)

\bibitem{hochreiter1998vanishing}
Hochreiter, S.: The vanishing gradient problem during learning recurrent neural
  nets and problem solutions. International Journal of Uncertainty, Fuzziness
  and Knowledge-Based Systems  \textbf{6}(02),  107--116 (1998)

\bibitem{kingma2014adam}
Kingma, D.P., Ba, J.: Adam: A method for stochastic optimization. arXiv
  preprint arXiv:1412.6980  (2014)

\bibitem{kingma2013auto}
Kingma, D.P., Welling, M.: Auto-encoding variational bayes. In: International
  Conference of Machine Learning (2013)

\bibitem{kingma2018glow}
Kingma, D.P., Dhariwal, P.: Glow: Generative flow with invertible 1x1
  convolutions. In: Advances in neural information processing systems. pp.
  10215--10224 (2018)

\bibitem{kraskov2004estimating}
Kraskov, A., St{\"o}gbauer, H., Grassberger, P.: Estimating mutual information.
  Physical review E  \textbf{69}(6),  066138 (2004)

\bibitem{krizhevsky2012imagenet}
Krizhevsky, A., Sutskever, I., Hinton, G.E.: Imagenet classification with deep
  convolutional neural networks. In: Advances in neural information processing
  systems. pp. 1097--1105 (2012)

\bibitem{liu2019coherent}
Liu, H., Jiang, B., Xiao, Y., Yang, C.: Coherent semantic attention for image
  inpainting. In: Proceedings of the IEEE International Conference on Computer
  Vision. pp. 4170--4179 (2019)

\bibitem{liu2019downscaling}
Liu, X., Lu, W., Zhang, Q., Huang, J., Shi, Y.Q.: Downscaling factor estimation
  on pre-jpeg compressed images. IEEE Transactions on Circuits and Systems for
  Video Technology  \textbf{30}(3),  618--631 (2019)

\bibitem{liu2020gradnet}
Liu, Y., Anwar, S., Zheng, L., Tian, Q.: Gradnet image denoising. In:
  Proceedings of the IEEE/CVF Conference on Computer Vision and Pattern
  Recognition Workshops. pp. 508--509 (2020)

\bibitem{liu2018large}
Liu, Z., Luo, P., Wang, X., Tang, X.: Large-scale celebfaces attributes
  (celeba) dataset. Retrieved August  \textbf{15}, ~2018 (2018)

\bibitem{mao2016image}
Mao, X., Shen, C., Yang, Y.B.: Image restoration using very deep convolutional
  encoder-decoder networks with symmetric skip connections. In: Advances in
  neural information processing systems. pp. 2802--2810 (2016)

\bibitem{nilsback2008automated}
Nilsback, M.E., Zisserman, A.: Automated flower classification over a large
  number of classes. In: 2008 Sixth Indian Conference on Computer Vision,
  Graphics \& Image Processing. pp. 722--729. IEEE (2008)

\bibitem{qin2019rethinking}
Qin, Z., Kim, D.: Rethinking softmax with cross-entropy: Neural network
  classifier as mutual information estimator. arXiv preprint arXiv:1911.10688
  (2019)

\bibitem{ronneberger2015u}
Ronneberger, O., Fischer, P., Brox, T.: U-net: Convolutional networks for
  biomedical image segmentation. In: International Conference on Medical image
  computing and computer-assisted intervention. pp. 234--241. Springer (2015)

\bibitem{saxe2019information}
Saxe, A.M., Bansal, Y., Dapello, J., Advani, M., Kolchinsky, A., Tracey, B.D.,
  Cox, D.D.: On the information bottleneck theory of deep learning. Journal of
  Statistical Mechanics: Theory and Experiment  \textbf{2019}(12),  124020
  (2019)

\bibitem{simonyan2014very}
Simonyan, K., Zisserman, A.: Very deep convolutional networks for large-scale
  image recognition. arXiv preprint arXiv:1409.1556  (2014)

\bibitem{srivastava2015highway}
Srivastava, R.K., Greff, K., Schmidhuber, J.: Highway networks. arXiv preprint
  arXiv:1505.00387  (2015)

\bibitem{szegedy2015going}
Szegedy, C., Liu, W., Jia, Y., Sermanet, P., Reed, S., Anguelov, D., Erhan, D.,
  Vanhoucke, V., Rabinovich, A.: Going deeper with convolutions. In:
  Proceedings of the IEEE conference on computer vision and pattern
  recognition. pp.~1--9 (2015)

\bibitem{thangaraj2017capacity}
Thangaraj, A., Kramer, G., B{\"o}cherer, G.: Capacity bounds for discrete-time,
  amplitude-constrained, additive white gaussian noise channels. IEEE
  Transactions on Information Theory  \textbf{63}(7),  4172--4182 (2017)

\bibitem{wah2011caltech}
Wah, C., Branson, S., Welinder, P., Perona, P., Belongie, S.: The caltech-ucsd
  birds-200-2011 dataset  (2011)

\bibitem{yan2018shift}
Yan, Z., Li, X., Li, M., Zuo, W., Shan, S.: Shift-net: Image inpainting via
  deep feature rearrangement. In: Proceedings of the European conference on
  computer vision (ECCV). pp. 1--17 (2018)

\bibitem{yu2018generative}
Yu, J., Lin, Z., Yang, J., Shen, X., Lu, X., Huang, T.S.: Generative image
  inpainting with contextual attention. In: Proceedings of the IEEE conference
  on computer vision and pattern recognition. pp. 5505--5514 (2018)

\bibitem{zhang2017beyond}
Zhang, K., Zuo, W., Chen, Y., Meng, D., Zhang, L.: Beyond a gaussian denoiser:
  Residual learning of deep cnn for image denoising. IEEE Transactions on Image
  Processing  \textbf{26}(7),  3142--3155 (2017)

\bibitem{zhang2018ffdnet}
Zhang, K., Zuo, W., Zhang, L.: Ffdnet: Toward a fast and flexible solution for
  cnn-based image denoising. IEEE Transactions on Image Processing
  \textbf{27}(9),  4608--4622 (2018)

\bibitem{zhang2020residual}
Zhang, Y., Tian, Y., Kong, Y., Zhong, B., Fu, Y.: Residual dense network for
  image restoration. IEEE Transactions on Pattern Analysis and Machine
  Intelligence  (2020)

\end{thebibliography}

}

% \newpage
% \onecolumn
% \appendix
% \input{secs/supplementary.tex}

\end{document}